%% file: main.tex
\documentclass[letterpaper, 10 pt, conference]{ieeeconf}  

\usepackage{times}
\usepackage{epsfig}
\usepackage{graphicx}
\usepackage{amsmath}
\usepackage{amssymb}
\usepackage{color}

\usepackage{multirow}
\usepackage{diagbox}

\usepackage{xspace}
\newcommand*{\eg}{e.g.\@\xspace}
\newcommand*{\ie}{i.e.\@\xspace}

\IEEEoverridecommandlockouts                          
\title{\LARGE \bf Learning Deep Visual Object Models From Noisy Web Data: \\ How to Make it Work}

\author{Nizar Massouh$^{1}$, Francesca Babiloni$^{1}$, Tatiana Tommasi$^{1}$, Jay Young$^{2}$, Nick Hawes$^{2}$ and Barbara Caputo$^{1}$
\thanks{This work was partially supported by the ERC grant 637076 - RoboExNovo (BC, TT), EU FP7 600623 STRANDS (JY, NH) and   the CHIST-ERA project ALOOF (BC, NM, JY, NH).}
\thanks{$^{1}$  Department of Computer, Control and Management Engineering,
        Sapienza Rome University, 00185 Rome, Italy
        {\tt\small \{massouh, babiloni, tommasi, caputo \} @dis.uniroma1.it}}%
\thanks{$^{2}$  Intelligent Robotics Lab, University of Birmingham, United Kingdom
        {\tt\small \{j.young, n.a.hawes \} @cs.bham.ac.uk}}%
}

\begin{document}

\maketitle

\thispagestyle{empty}
\pagestyle{empty}

\begin{abstract}
\input{abstract}
\end{abstract}

\section{Introduction}
\input{intro}

\section{Related Work}
\label{related}
\input{related}

\section{Studying the effect of noisy labels on CNNs}
\label{study-noise}
\input{study-noise}

\section{Learning from the Web}
\label{web_learning}
\input{web_learning.tex}

\section{Experiments}
\label{robot}
\input{robot}

\section{Conclusions}
\label{conclu}
The aim of this work is to pave the way for robots able to learn perceptual object models from images found on the Web, without the need of human annotators. We took the ImageNet ILSVRC12 1000 categories database as a reference task, and we studied how injecting noise into it affected the performance of two popular deep architectures, AlexNet and GoogLeNet. We built on our findings, presenting a query expansion approach for automatically re-creating ILSVRC12 from Web downloads, and we demonstrated experimentally that this Web-derived version of ImageNet leads to deep models that are better at capturing the visual facets of object categories when used 
to extend a robot on-board perceptual object knowledge.

We see this work as the first step on a long and intriguing road. The possibility to  automatically create data collections from the Web, large enough to train a deep net, should be further explored, aiming for much larger data collections in terms of classes and images per class. This would allow us to study questions such as when would we reach the upper bound of accuracy for a given architecture? Another avenue we intend to explore is the possibility of creating task specific, rather than all-purpose, databases, which intuitively should allow a robot to exploit any prior knowledge it might have on its situated settings and goals. Finally, we plan to continue our investigations on the effect of noise on the behavior of deep architectures, both in terms of further experiments and in-depth analysis trying to pinpoint the specific layers that are more affected by it.

\addtolength{\textheight}{-12cm}  

\bibliographystyle{IEEEtran}
\bibliography{main.bbl}

\end{document}

%% file: abstract.tex
Deep networks 
thrive when trained on large scale data collections. This has given ImageNet a central role in the development of deep architectures for visual object classification.  
However, ImageNet was created during a specific period in time, and as such it is prone to aging, as well as dataset bias issues.
Moving beyond fixed training datasets will lead to more robust visual systems, especially when deployed on robots in new environments which must train on the objects they encounter there.
To make this possible, it is important to break free from the need for manual annotators. Recent work has begun to investigate how to use  the massive amount of images available on the Web in place of manual image annotations.
We contribute to this research thread with two findings: (1) a study correlating a given level of noisily labels to the expected drop in accuracy, for two deep architectures, on two different types of noise, that clearly identifies GoogLeNet as a suitable architecture for learning from Web data; (2) a recipe for the creation of Web datasets with minimal noise and maximum visual variability,
 based on a visual and natural language processing concept expansion strategy. 
 By combining these two results, we obtain a method for learning powerful deep object models automatically from the Web.
We confirm the effectiveness of our approach through object categorization experiments using our Web-derived version of ImageNet on a popular robot vision benchmark database, and on a lifelong object discovery task on a mobile robot.

%% file: intro.tex
\begin{figure*}[h]
\hspace{-5mm}
\begin{tabular}{cccc}
\includegraphics[width=0.24\linewidth]{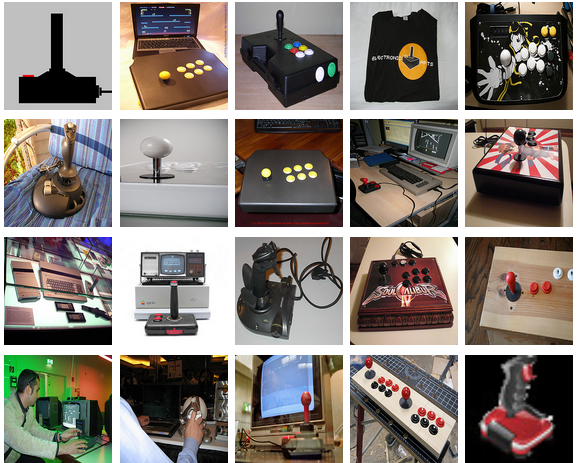} &
\includegraphics[width=0.24\linewidth]{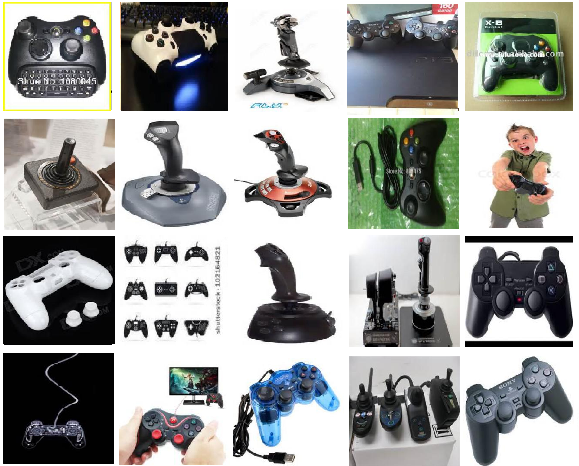} &
\includegraphics[width=0.24\linewidth]{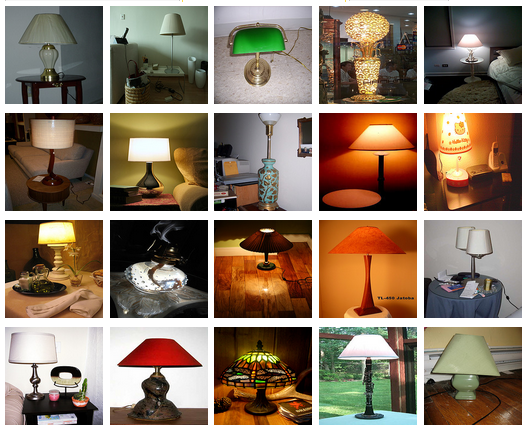} &
\includegraphics[width=0.24\linewidth]{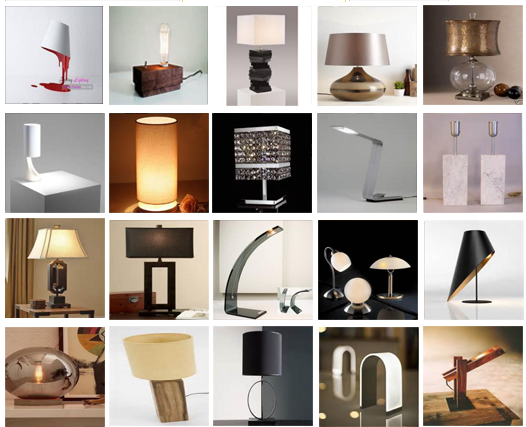} \\
Joystick - ILSVRC12 (IN)&
Joystick - WINC &
Table Lamp - ILSVRC12 (IN)&
Table Lamp - WINC \\
\end{tabular}
\caption{Exemplar images for the classes ``Joystick'' and ``Table Lamp'' for ImageNet ILSVRC12 (IN) and its replica created automatically from the Web (WINC). We see the evident differences in style for both classes, brought by time.}
\vspace{-3mm}
\label{figure:imagenet}
\end{figure*}

Deep networks and ImageNet are the key ingredients of the most recent and successful object categorization approaches in robot vision today, across 2D \cite{Behnke,jhuit}, 2,5D \cite{gupta2014learning,eitel2015multimodal} and 3D \cite{socher2012convolutional,zaki2016convolutional} data. 
The progress achieved by learning methods in object classification over the last five years has been impressive, and it goes hand-in-hand with the development of deep architectures. In turn, the success obtained so far relies heavily on the availability of very large scale annotated data collections. 
ImageNet \cite{deng2009imagenet}, and to a lesser extent Places \cite{places2014}, are, at the time of writing, the only two publicly available resources suitable for training deep learning-based object categorization algorithms. The overwhelming majority of such systems use these two collections for pre-training, usually followed by some form of network adaptation to a specific task  (for a review of the relevant literature see Section \ref{related}).

As with any other annotated database, the perceptual object knowledge in ImageNet is \emph{static}, i.e. its main body was collected during a limited and specific period (circa 2010\footnote{http://image-net.org/about-stats.}). This can result in some classes becoming dated over time, especially those representing man-made objects (see Figure \ref{figure:imagenet}). 
Robots deployed in everyday environments, such as offices, homes or hospitals, will encounter a wide variety of man-made objects. Being trained on out of date models will reduce their performance. Moreover, large-scale image datasets are inevitably collected by multiple annotators, which might involuntarily inject some of their own views and bias on object categories, resulting in dataset bias \cite{DBLP:conf/cvpr/TorralbaE11}. 
Fine-tuning, \ie adapting a pre-trained architecture to a new classification task, requires annotated data, computational resources and computing time. For several robotic settings, these might not be feasible options.
%



The awareness that we need more diverse data collections to feed data-hungry deep architectures is growing in the community. A promising research thread is the attempt to replace real images with synthetically generated data \cite{Carlucci}. 
An alternative approach attempts to create large-scale databases by downloading images directly from Web search engines \cite{cheng2015semantically}, without any subsequent cleaning by human annotators.  Here the main challenge is to cope with noisy (automatic) annotations \cite{DBLP:journals/corr/SukhbaatarF14}, while at the same time making sure that the collected images  are sufficiently representative of all the visual facets of a given object class.

This paper contributes to this latter approach. We first argue that to take full advantage of noisily labeled data, one needs to study the effect of noise on labels in controlled settings. To this end,  
we inject different percentages of noise, of various kinds, into the labels of ImageNet, and study its impact on accuracy prediction in two different deep architectures.
%
%
Our study shows that while the accuracy of the popular AlexNet architecture \cite{krizhevsky2012imagenet} degrades smoothly as the level of noise increases, GoogLeNet \cite{googlenet} maintains a level of accuracy very close to its top performance for levels of noise up to $35\%$, then it experiences very severe convergence issues, de facto preventing its use for classification.  This finding provides an indirect way to assess the level of noise in a Web-generated database, without any need for manual annotation: if GoogLeNet converges, then the noise level in the automatically generated data collection will be between $0\%$ and $35\%$, and it will have a minimal impact on the accuracy achievable by the architecture on those data. This result is the first contribution of this paper.

Our second contribution is
an algorithm for the automatic creation of large data collections from the Web that minimizes noise and maximizes image variability
through visual and natural language processing, performing a \emph{concept expansion} of the query.
We use our approach to re-create automatically ImageNet from the Web and we compare what we get against a dataset collected
without query expansion, performing a qualitative evaluation in terms of visible label noise and a quantitative analysis in terms of
CNN classification accuracy when using our two deep nets of choice.

We further demonstrate the effectiveness of our approach in
two robotics scenarios. We show that CNN networks trained on Web-collected data can be used as feature extractors
for images of a popular robot object database \cite{washington} and allows to get on it a classification accuracy
on par to that obtained when using ImageNet-trained CNN models. Moreover, for a robot deployed in an indoor environment
and able to use deep models to overcome gaps into its own visual knowledge base without human intervention,
our Web-generated models provide a significant increase in performance compared to using the original ImageNet.

The rest of the paper is organized as follows: after a review of the relevant literature (section \ref{related}), section \ref{study-noise} describes the protocol developed to study the effect of noisy labels on deep networks, and reports our findings. Section \ref{web_learning} describes our query conceptual expansion strategy, illustrating the impact of each component with an experimental study. Section \ref{robot} showcases the power of our approach on object categorization and robotic scenarios. We conclude identifying open issues and future research directions we plan to pursue.

%% file: related.tex
Since the seminal work of Krizhevsky \cite{krizhevsky2012imagenet}, deep learning has significantly changed the research landscape in visual object recognition. Over the last few years, the use of convolutional neural networks trained over large-scale databases has become the cornerstone of most robot visual systems. 
An important question is how to tailor the performance of such networks to the specific task at hand. Deep networks need large scale annotated databases during training. To this end, ImageNet and particularly the related Large Scale Visual Recognition Challenge subset (ILSVRC12, \cite{deng2009imagenet}) are widely used in combination with various deep architectures \cite{Behnke,Carlucci,zaki2016convolutional,eitel2015multimodal}. To adapt such 1000-categories classifiers to a new problem, a common strategy is fine-tuning \cite{DBLP:conf/bmvc/ChatfieldSVZ14}, i.e. the procedure of re-training a part of the network using annotated data from the new classification problem of interest, while keeping the parameters in other parts of the network fixed. The array of fine-tuning strategies is very large \cite{DBLP:conf/bmvc/ChatfieldSVZ14,Behnke}, and although effective, the approach depends strongly on heuristics. Moreover, the assumption of having an annotated dataset of $10^3$ images from the new task might be unrealistic in several robotics domains, where a system should be able to perceive and act without lengthy training on the site, that would be in any case static as well. 

The idea of looking at the Web as a source of knowledge from which to learn is not new in robotics. Previous work looked at obtaining semantic information from the Web in various forms, \eg from structured and semi-structured sources like WikiPedia, DBPedia and WordNet \cite{DBLP:conf/icra/BeetzBWWBBBBFFH16,12aaai-SamadiKollarVeloso,young_ecai16}, or combined with deep networks pre-trained on ImageNet \cite{Young2017}. Mining the Web to learn about the appearance of objects or scenes is much less investigated. Ventura \emph{et al}  proposed using the Web to call upon human annotators if a robot detected  gaps in its perceptual knowledge \cite{13aaaiw-VenturaColtinVeloso}. Wohlkinger \emph{et al} downloaded 3D CAD models from public Web resources, creating a 3D object database acting as a proxy for the Web \cite{DBLP:conf/icra/WohlkingerARV12}. Recent work by Young \emph{et al} \cite{Young2017} proposed the combination of semantic and perceptual Web knowledge, using curated resources like DBPedia and ImageNet as Web proxies. We are not aware of previous work in robot vision for automatically mining the Web for RGB images, in order to learn perceptual object models on a robot without use of manual annotators. 

The automatic creation of image databases by mining the Web has been researched in the computer vision community. Focusing explicitly on the work done within the context of deep network training, research efforts have followed two different directions. The first acknowledges the fact that databases created from the Web without manual intervention will inevitably contain some level of noise,  and so attempt to design loss layers able to cope with this issue \cite{DBLP:conf/iccv/ChenG15,DBLP:journals/corr/SukhbaatarF14}. The second tackles this issue by proposing query expansion strategies that should reduce the level of noisily annotated images while preserving the visual variability necessary to successfully train deep nets \cite{cheng2015semantically,DBLP:conf/cvpr/DivvalaFG14,DBLP:conf/eccv/GolgeD14}. Our work sits in the middle of these two directions: as in the first, we assume that images downloaded automatically from the Web will have some degree of noise associated with their label. Rather than designing regularization layers to cope with this,  we instead study the effect of noisy labels on two different deep architectures and use the insights of this study to our advantage. This, combined with a new recipe for automatic query expansion able to reduce noisily annotated data while spanning the visual concept space significantly, lead to a powerful strategy for robot visual classification in the wild.

%% file: study-noise.tex
When automatically harvesting the Web to build databases for object classification, the labels of the images can be expected to contain noise. However, the level 
of noise for any given strategy is seldom investigated, beyond the manual generation of ground truth. When building databases of the scale necessary for deep learning, assessing this noise directly is a daunting task.

We therefore assess the level of noise associated with labels created by a given automatic database collection procedure through a reference task. 
We consider ImageNet, specifically its ILSVRC12 set of $1000$ categories, as our classification problem. 
We evaluate how the accuracy 
of different deep architectures changes when the original dataset is injected with different levels and types of noise. Assessing such changes in classification accuracy gives us information about how well 
networks trained from scratch on general purpose databases will perform on new, Web-collected data without human annotation. 

\begin{figure*}[t!]
\begin{center}
    \includegraphics[width=0.95\linewidth]{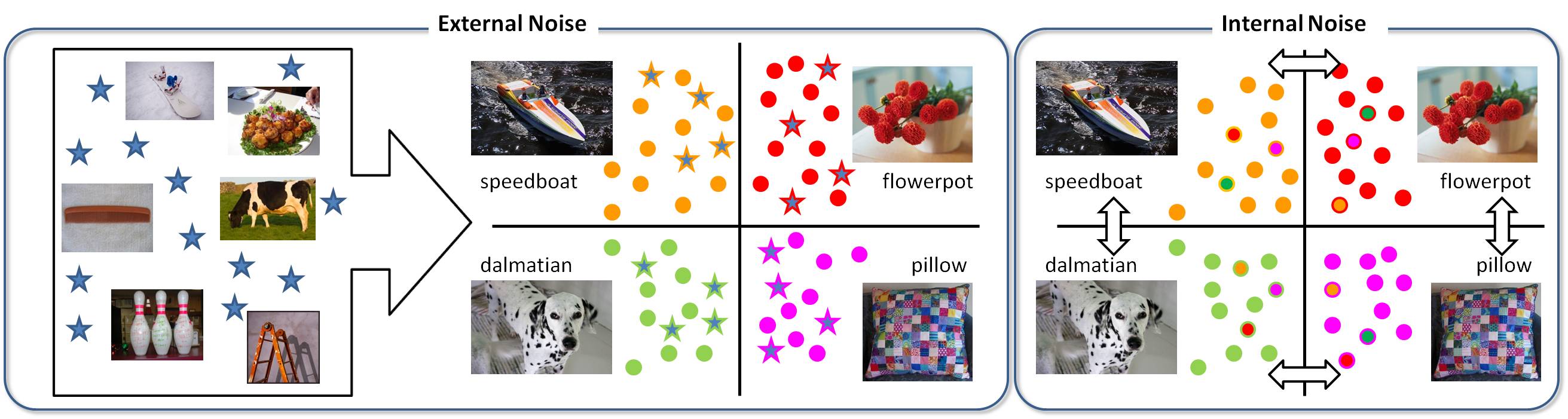} 
\caption{Schematic illustration of the procedure used to inject different kind of noise into the original
ILSVRC12 data collection. Left: in the external noise case, ImageNet images of categories not involved in the
ILSVRC12 set are labelled with the name of the ILSVRC12 categories and substituted to various percentage of the 
original images. Right: in the internal noise case, some of the images have their label switched to that
of another class in the set.}\vspace{-3mm}
\label{fig:scheme_noise}
\end{center}
\end{figure*}

We focus on two main types of label noise: 
\begin{itemize}
\item \emph{external} noise describes an image that is associated with a label, but does not actually belong to any label in the dataset;
\item \emph{internal} noise describes an image that is associated with a label, but actually belongs to another label in the dataset. 
\end{itemize}
We created replicas of the ILSVRC12 collection with increasing
percentages of noise. We created separate replicas for internal and external label noise, ranging from $5\%$ to $85\%$ in increments of $10\%$. 
We distributed erroneously labeled images uniformly  over the classes, substituting 
the original correct images with samples pooled randomly either from other ImageNet 
classes not belonging to ILSVRC12 (external noise) or from different classes of the same set 
(internal noise). This is illustrated in Figure \ref{fig:scheme_noise}.

We tested two different CNN architectures to evaluate their robustness to this type of noise: \emph{AlexNet} \cite{alexnet} 
and \emph{GoogLeNet} \cite{googlenet}. 
The first is the na\"ive standard CNN in object categorization, while the second is a deeper architecture which 
includes Inception modules designed to improve the network effectiveness without an uncontrolled blow-up 
of the computational complexity. In all our experiments we trained the networks on the noisy sets,  
evaluating the classification performance on the original ILSVRC12 validation set. We used the Caffe \cite{jia2014caffe} CNN
implementation available on NVIDIA Deep Learning GPU Training System
(DIGITS) for all our experiments. The learning rate was set to $0.01$ initially and divided by 10 after every third of the total epochs. We trained both networks with an SGD solver for 30 epochs and we repeated each experiments 5 times reshuffling the image list and using batch size of 128 for AlexNet and 32 for GoogLeNet. 
\begin{figure}[t!]
\begin{center}
\includegraphics[width=0.95\linewidth]{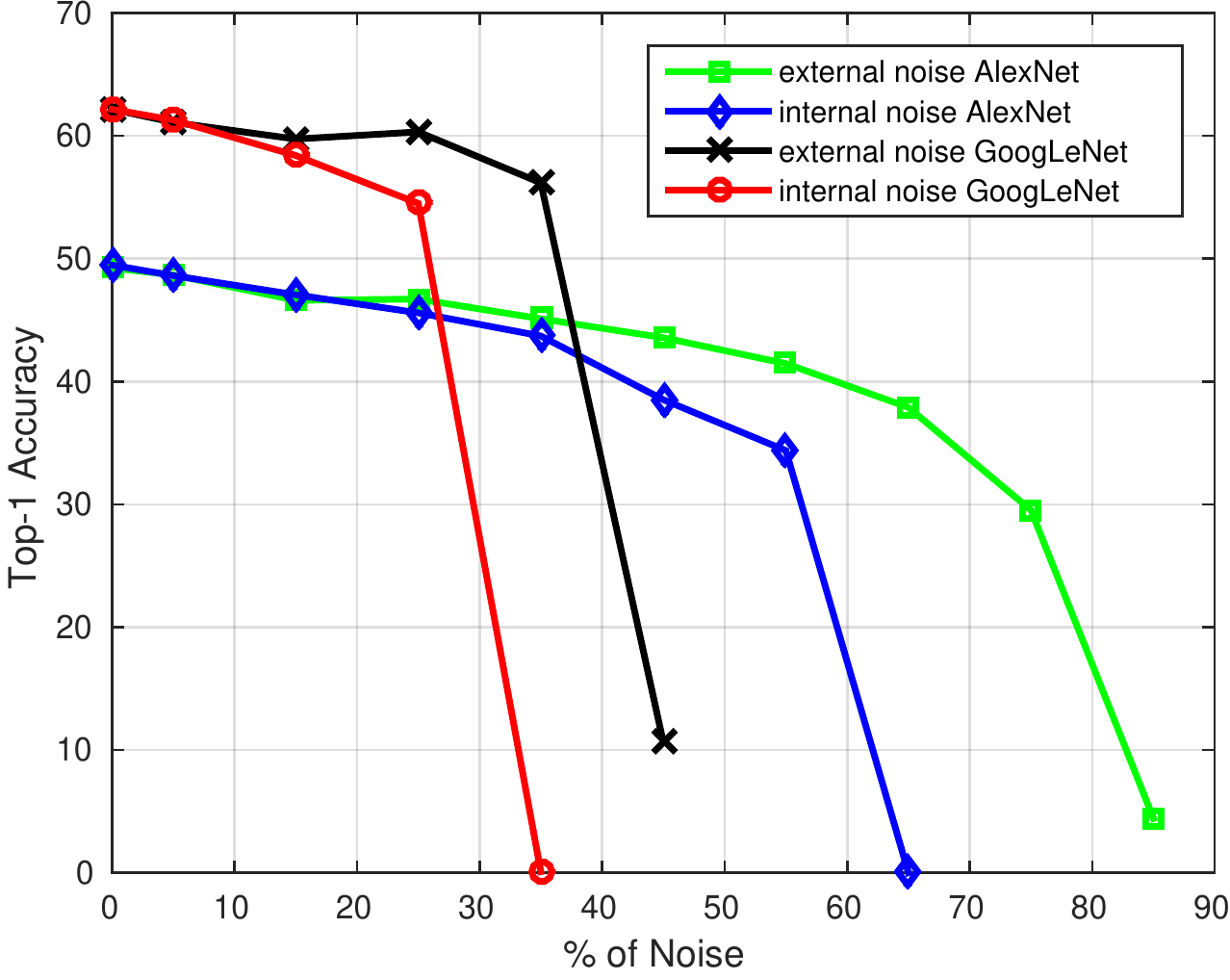} 
\caption{Top-1 classification accuracy produced by networks trained on ILSVRC12 data containing different
percentage of noise and evaluated on the original ILSVRC12 validation set.}
\label{fig:plot_noise}\vspace{-3mm}
\end{center}
\end{figure}
The average accuracy results are shown in Figure \ref{fig:plot_noise}. We see that 
a low percentage of internal or external noise ($<20\%$) induces only a moderate reduction 
in classification performance. For larger amounts of noise ($\ge 20\%$), the networks
appear more robust to external noise than to internal noise. Moreover, GoogLeNet outperforms
AlexNet for a low noise percentage, while the trend inverts for an amount of noise equal or over 35\%. 
Although the former result is expected \cite{googlenet}, the latter is quite surprising, 
 indicating that GoogLeNet, besides being deeper, is also less robust to noise than AlexNet.

A further investigation conducted on the external noise case indicated that for 35\% noise, 
GoogLeNet  converged during training in three out of the five repetitions of the experiment, 
while for 45\% noise it converged in only one of the runs. Since the repetitions only 
differed in the data order, this tells us that the content of each data batch fed to GoogLeNet 
is important.  When every batch is noisy it becomes difficult to optimize the 
large number of network parameters. We also tried doubling the batch size 
with no significant difference in the overall result trend.

To the best of our knowledge, this GoogLeNet behavior has not been observed or discussed
before. Besides drawing attention to a limitation of very deep CNN architectures, 
this result suggests that GoogLeNet might be used as a rough indicator of the level of label noise present in an image database generated automatically from the Web: if for a given collection GoogLeNet converges, this means that (a) the level of noise on the labels is at most $40\%$, and (b) the recognition accuracy using this training data is expected to be roughly $5\%$ lower than would be achieved by using a noise-free version of the same database. In the rest of the paper we will show how this result, coupled with an effective approach for automatic Web mining, leads to state of the art results in lifelong object discovery \cite{Young2017}.

%
%

%% file: web_learning.tex
This section illustrates our strategy for collecting images from the Web in order to build a training set for learning deep classification models. We present a qualitative analysis on the gathered data and a quantitative evaluation on the accuracy obtained when using them to train deep nets. For a reference task we focus on reproducing the ILSVRC12 database of 1000 categories \cite{ILSVRC15}. 

\subsection{Dataset Collection}
Collecting images of a given category from the Web presents some 
practical challenges. Many popular search engines have 
restrictions that limit downloads to a maximum of 1000 images per query. Considering the large number of duplicates for some categories, 
the  
number of images obtained per query can be very limited. This limitation means that the statistics of
the original ILSVRC12 dataset, composed by several categories each represented by $10^3$ images,  are hard to recreate with direct query methods. 

We can tackle this issue with two possible solutions: 
\begin{itemize}
\item \emph{category name \& single search engine} - use the category name to query a search 
engine that has weak download restrictions and provides a large number of low quality images;
\item \emph{query expansion \& multiple search engines} - expand the query by adding extra textual 
information to the category name and use multiple search engines, each with strict download limitations.
\end{itemize}
The first strategy can use search engines like
Picsearch (\emph{www.picsearch.com}) from which we can get more than 7000 images per class, but the
level of control on the output is quite limited (\eg it is not possible to choose the 
image type to exclude drawings), and the relevance of the images with respect to the query 
decreases quickly. The second strategy needs external 
expert knowledge for each category to provide related text and expand the original query.

One widely adopted method to obtain meaningful expansions is to leverage WordNet \cite{wordnet}, 
either considering the synonyms in the synset of each category, the parent nodes, 
or even the whole category textual description.
However, some of the WordNet synsets contain a single word (\eg ``jay'', ``boxer'', ``crane'') so no synonyms are
available. When searching for parents, deciding at which depth level of the ontology to stop
is not trivial (\eg ``jay'', ``corvin'', ``oscine'', ``passerine'', ``bird'', ``vertebrate'', etc.): 
the hierarchy of classes is not uniform, with some parts more or less dense than others due to different 
numbers of siblings.
Hence, it is hard to find a  meaningful rule for all categories.
Using the textual description of a category requires a natural language pre-processing step
able to isolate relevant words and avoid the confusing ones (\eg for "crane": ``large long-necked 
wading bird of marshes and plains in many parts of the world'' the query \emph{crane+bird} can be useful,
but \emph{crane+world} provides misleading results). 
Finally, to get a sufficient amount of images, one or two words are not enough, as preliminary tests show that 
more than ten are needed to reach our goal of reproducing the ILSVRC12 class statistics.

In the following we will use \emph{IN} to indicate the original ImageNet ILSVRC12 dataset and 
\emph{WIN} to indicate our Web-collected ImageNet ILSVRC12 version, obtained adopting following the \emph{category name \& single search engine} strategy. 
We also create \emph{WINC} using \emph{query expansion \& multiple search engines} strategy. Below we describe our query expansion and preprocessing procedures.

\textbf{Query Expansion.}
Since there isn't a single universal strategy for query expansion, we target 
simplicity. 
We assume that our database creation process can only access the Web to gain knowledge and cannot ask for a human for disambiguation or expert help. 
Under such conditions, Search Engine Optimization (SEO) practice can provide guidance. 
SEO evaluates the relation between how users query a search engine and the 
preference they express in the returned results. A highly preferred result, usually 
top ranked by the search engine, is indexed with \emph{keywords} considered as very relevant to 
the original query. By exploiting SEO tools, in particular the ones that target image
search, we can easily get a large number of relevant keywords associated to visual concepts.
Here we use the MyKeyworder service (\emph{www.mykeyworder.com}), giving
a category name as input and collecting the top 20 output keywords. These can be
considered intrinsically visual, because they are associated with image search queries, semantic concepts and 
photographic tags. Each of the obtained keywords are then used as query expansions for an object 
class name when querying three image search engines: Google, Yahoo! and Bing. Although the maximum 
number of available images from each of them is 1000, we limited ourself to the top 200 results as 
 in \cite{Chatfield12}. This choice reduces the risk of introducing irrelevant 
and noisy images (which usually appears after the first result page), and guarantees a large 
variability in the final image set since multiple keywords cover different visual aspects 
of the same concept. Considering the whole set of 1000 ILSVRC12 classes, the download phase leaves us with 
$20 \times 3 \times 200 \times 1000 = 12 \times 10^6$ images and a large space for further refinement. 

\textbf{Duplicates Removal.} An essential preprocessing step consists of removing duplicate images. 
When using multiple search engine as well as multiple concept expansions, we deal with two kinds of 
image repetitions. The \emph{per-query duplicates} are exact copies or very similar images obtained
by using the same query across three different search engines. These duplicates are eliminated immediately
after the end of the download process. The \emph{cross-expansion duplicates} are exact copies or very 
similar images obtained when expanding a concept with different keywords. These duplicates are removed
in a second processing step; the number of images shared between each expansion and the original 
 concept, not expanded, is saved as a measure of diversity. 
Both the duplicate removal processes are executed by extracting the image fingerprints through perceptual 
hashing and then comparing them to find close image pairs.

\begin{figure}
\includegraphics[width=1\linewidth]{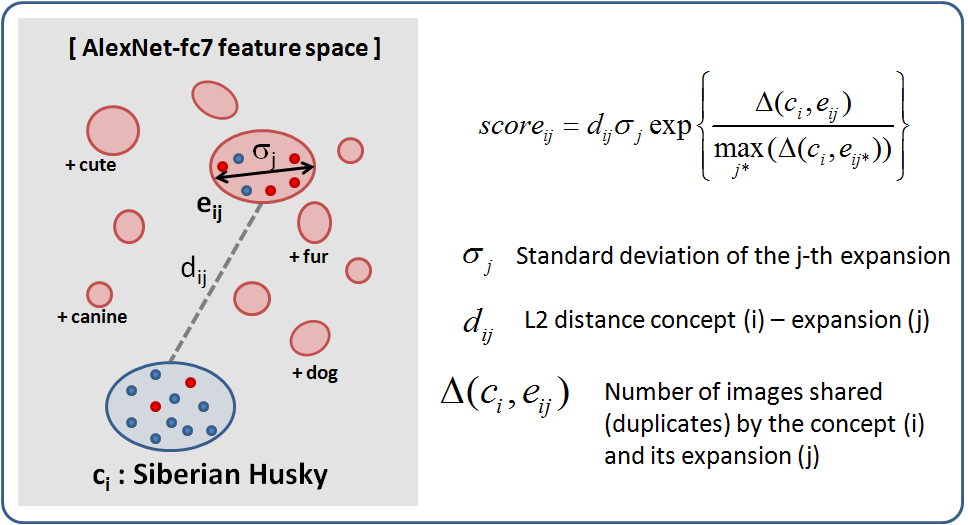}
\caption{Schematic illustration of the relation between a concept and its expansions in the feature
space. By using the distribution of the samples we define a score for each expansions which helps 
to select them by maximizing the variability coverage for each class. 
In our experiments we used off-the-shelf CNN fc7 activations of 
an AlexNet pre-trained on ImageNet ILSVRC12  as features \cite{Off-the-Shelf}. }
\label{fig:formula}
\end{figure}

\textbf{Variability Maximization.}
To induce better generalization in the classifier that will be learned on this data we aim to maximize the quality
and variability of each class. Therefore we select only high quality expansions to include in the dataset. Given the $j$-th expansion $e_{ij}$ of the $i$-th class $c_i$, we can reasonably suppose that it is of high quality if,  
\begin{itemize}
\item  the images of the expansion $e_{ij}$ lie close to the class images $c_i$, \ie they have a small L2 distance $d_{ij}$ in the feature space;
\item  the images of the expansions $e_{ij}$ are not sparse, i.e. there is small standard deviation $\sigma_{j}$ in the image set; 
\item  the number of duplicates $\Delta$ shared by the pair $(c_i, e_{ij})$ is small, which ensures variability.
\end{itemize}
We define an inverted score for each expansion (minimal score corresponds to maximal quality, see Figure 
\ref{fig:formula} for the formula and a graphical illustration), sort them, pick the top 5 expansions out of the 20 per
class and merge the corresponding images to define the pool from which we randomly select samples with the same statistics of the ILSVRC12 collection.
Table \ref{table:expansion} shows the effect of the query expansion and of the variability maximization 
procedure, indicating the extra keywords obtained with the former and the selection operated by the latter for four categories.
\begin{table*}[h]
\centering
\caption{For the four categories listed in the first column, this table shows the top 20 keywords produced as output
by the SEO tool (second column) and the 5 keywords selected by the variability maximization procedure (third column).}
\vspace{-2mm}
\begin{tabular}{|c|c|c|}
\hline
class name & keywords for concept expansion & selected concepts \\\hline
\multirow{2}{*}{siberian husky} & 
dog, siberian, animal, canine, cute, portrait, pet, breed, mammal, purebred, & \multirow{2}{*}{pet,isolated,beautiful,purebred,dog}\\
& domestic, young, white, fur, beautiful, nature, wolf, isolated, adorable, looking &\\ \hline
\multirow{2}{*}{grey whale} & 
mammal, ocean, sea, wildlife, marine, mexico, gray, animal, life, water, & \multirow{2}{*}{mexico,ocean,swimming,lagoon,water}\\
&lagoon, cetacean, fin, baja, blue, swimming, tail, watching, america, powerful &\\\hline
\multirow{2}{*}{desk} & office, top, table, work, business, background, white, view, blank, empty,
& \multirow{2}{*}{office, space, design, above, phone}\\
& above, space, computer, paper, workplace, wooden, coffee, phone, notebook, design &\\ \hline
\multirow{2}{*}{vending machine} & business, buy, dispenser, vector, automatic, drink, food, service, coin, isolated,
& \multirow{2}{*}{beverage, food, automatic, can, drink}\\
& illustration, consumer, can, beverage, snack, choice, symbol, sale, merchandise, button &\\ \hline
\end{tabular}
\label{table:expansion}
\end{table*}

\subsection{Qualitative Analysis}
\label{qualitative}
\input{qualitative}

\subsection{Quantitative Analysis}
\label{quantitative}
\input{quantitative}

%% file: qualitative.tex
\begin{figure*}[tb]
\hspace{-5mm}
\begin{tabular}{cccc}
\includegraphics[width=0.24\linewidth]{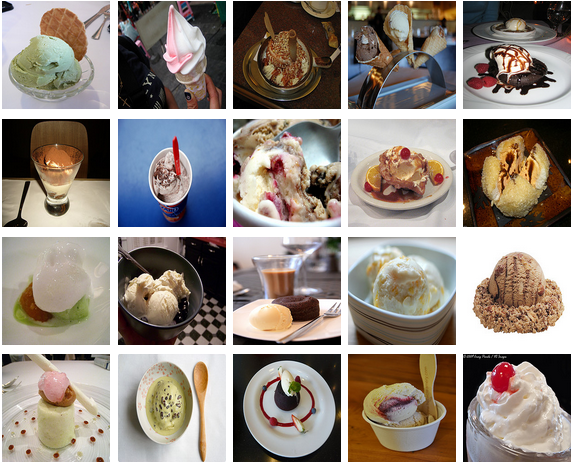} &
\includegraphics[width=0.24\linewidth]{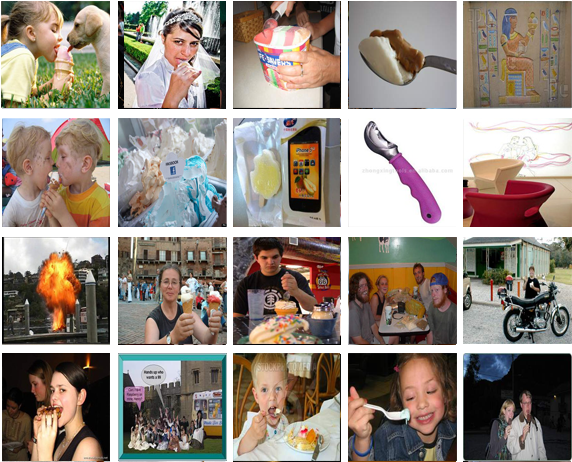} &
\includegraphics[width=0.24\linewidth]{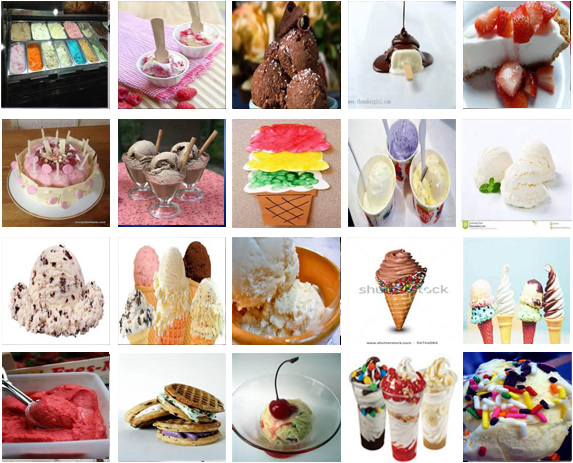} &
\includegraphics[width=0.26\linewidth]{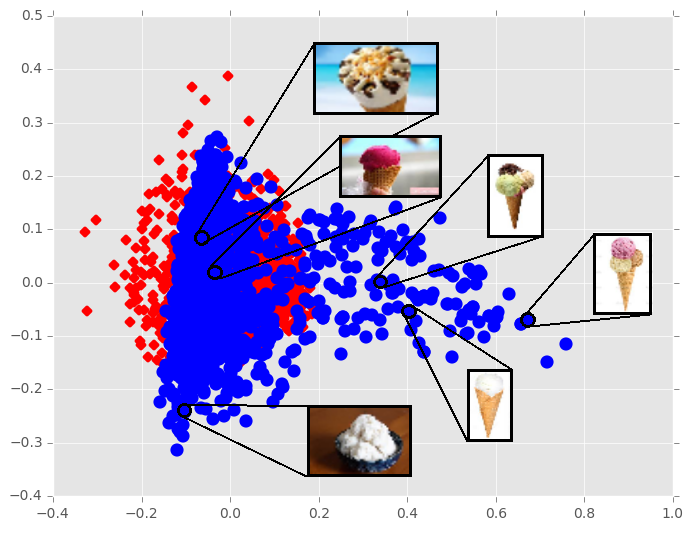} \\
Icecream - ILSVRC12 (IN)&
Icecream - WIN &
Icecream - WINC &
ILSVR red, WINC blue \\
\includegraphics[width=0.24\linewidth]{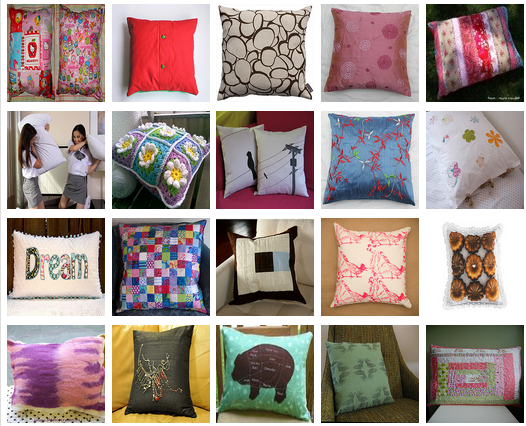} &
\includegraphics[width=0.24\linewidth]{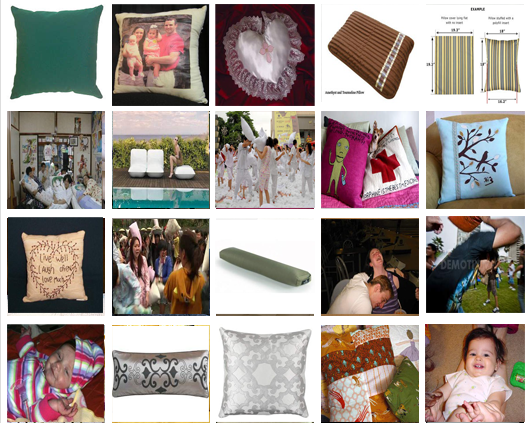} &
\includegraphics[width=0.24\linewidth]{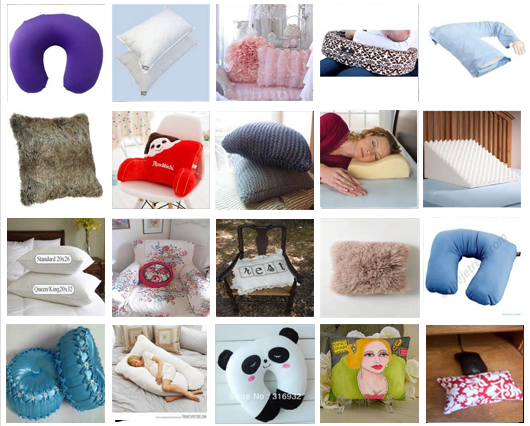} &
\includegraphics[width=0.28\linewidth]{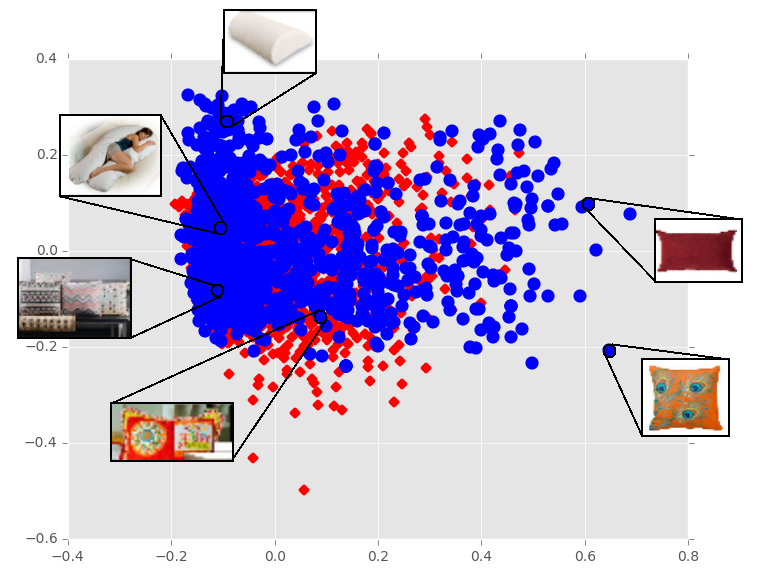}\\
Pillow - ILSVRC12 (IN) &
Pillow - WIN &
Pillow - WINC &
ILSVR red, WINC blue
\end{tabular}
\caption{Examples of images extracted from ImageNet ILSVRC12 (IN), WIN and WINC dataset for two categories. The last column on the
right shows dot plots of image samples for IN and WINC in a bi-dimensional subspace obtained applying PCA on the fc7 activations of AlexNet trained on ImageNet ILSVRC12.
}
\label{fig:IN-WIN-WINC}
\end{figure*}
To show the difference between the reference \emph{IN}  dataset and \emph{WIN/WINC}
we show samples of two object categories in Figure \ref{fig:IN-WIN-WINC}. For both ``Icecream'' and  ``Pillow''
we see that (1) \emph{WIN} images often do not focus on the single object and consider instead a whole 
scene; moreover, some of the images are irrelevant.
(2) \emph{WINC}  images appear much more consistent with the corresponding assigned label; 
(3) by projecting the feature maps (fc7 activations of AlexNet trained on ImageNet) of the images  in the 
same class of \emph{IN}  and \emph{WINC} into a bi-dimensional space through PCA
we can look at their overall distribution. The plots show that \emph{WINC}, besides spanning the same part of the 
visual space of \emph{IN}, presents also a larger variability which is not only due to noise,
but also to several relevant images (Icecream cones and Pillows of different shapes on white background).
Similar conclusions can be drawn when looking at other object categories.

%% file: quantitative.tex
\begin{table*}[h]
\centering
\caption{Top-1 accuracy performance ($\%$): classification results of two CNN architectures trained 
on the curated \emph{IN} dataset and on its web-collected replicas \emph{WIN} and \emph{WINC}.}
\vspace{-2mm}
\begin{tabular}{ccccccccccc}
\cline{2-6}\cline{8-11}
                                 & \multicolumn{5}{c}{val:IN} && \multicolumn{4}{c}{val:WINC}\\
\cline{1-6}\cline{8-11}
\backslashbox{train}{n. val. classes} &  10   & 50   & 100  & 200  & 1000 && 10   & 50   & 100  & 200 \\
\cline{1-6}\cline{8-11}
AlexNet IN  & 57.8 & 59.5 & 61.7 & 58.5 & 49.5 && 58.8 & 61.1 & 57.9 & 56.4\\
AlexNet WIN & 38.8 & 31.9 & 33.2 & 32.5 & 31.0 && 47.2 & 38.8 & 38.7 & 39.1\\
AlexNet WINC& 42.2 & 39.7 & 42.3 & 37.9 & 33.6 && 62.1 & 61.0 & 59.7 & 59.2 \\ 
\cline{1-6}\cline{8-11}

GoogLeNet IN & 64.8 & 65.8 & 66.0 & 65.3 & 62.2 && 63.4 & 65.0 & 62.1 & 61.4 \\
GoogLeNet WINC & 52.2 & 51.2 & 50.5 & 49.6 & 42.7 && 69.2 & 70.4 & 65.1 & 65.0 \\
\hline
\end{tabular}\vspace{-1mm}
\label{table:quantitative}
\end{table*}

%
%
%
%
%
%

We complete the evaluation of the Web-collected datasets by training CNNs on them 
and comparing their performance with that obtained  on 
the curated ImageNet ILSVRC12 (\emph{IN}). 
Specifically, we train AlexNet and GoogLeNet,  testing
both of them on the original ILSVRC12 validation set (\emph{val:IN}) as well as 
on a cleaned Web-based validation set (\emph{val:WINC}). This was collected by 
selecting 50 images out of the samples of \emph{WINC} that were not involved in the 
respective training set and manually verifying their labels. Although this manual image 
annotation is time expensive, having this extra testbed allows us to assess the effect 
of a possible bias between ImageNet and the Web-collected datasets as discussed in 
the introduction. To keep the time dedicated to the annotation limited, we considered a
subset of 200 out of 1000 classes. For completeness
we also ran tests on groups of 10, 50 and 100 classes, showing that the overall
performance of the different models remain consistent regardless of the cardinality 
of the labeled set.

The top-1 classification accuracy results obtained with AlexNet are shown in the first rows
of Table \ref{table:quantitative}. They indicate that
the models trained on \emph{IN} outperform those trained on \emph{WIN/WINC}  when both are tested 
on \emph{val:IN}. Between \emph{WINC}  and \emph{WIN} the first shows higher accuracy than the second, as we 
could expect from the qualitative analysis.
The trend between \emph{IN} and \emph{WINC}  changes when testing on \emph{val:WINC}. Here training on \emph{WINC}
provides equal or better performance than training on \emph{IN}. These results prove empirically 
the presence of a domain shift between the ImageNet and Web underlying probability distributions
\cite{Saenko:2010}. 
They also allow us to claim the reliability of the \emph{WINC} model.
The behavior of the \emph{WIN} model remains instead unchanged regardless of the validation set, 
indicating that this version of the Web-collected dataset is significantly worse than the
original \emph{IN}. Because of this, we will not consider it in the following analysis, concentrating 
only on our clean \emph{WINC}.

When using GoogLeNet (bottom part of Table \ref{table:quantitative}) all  
results grow in absolute terms, but the relative performance of the \emph{IN} and \emph{WINC} models 
remains the same as with AlexNet, confirming what already stated before. 
A further important message provided by the GoogleNet  
results is that it is possible to train this network on \emph{WINC} without encoutering network convergence issues. From our results in Section \ref{study-noise} we can conclude that the amount of 
noise present in our Web-collected version of ImageNet is below the critical threshold 
of $35\%$.

%
%
%
%

%% file: robot.tex
Having shown that it is possible to learn reasonable CNN classification models from
Web-collected images, we now assess their performance when facing real scenarios from robotic 
applications. We consider two tasks: object classification using the models for
feature extraction, and lifelong object discovery using the models to produce category proposals.
\begin{figure}[t]
\centering
\includegraphics[width=0.8\linewidth]{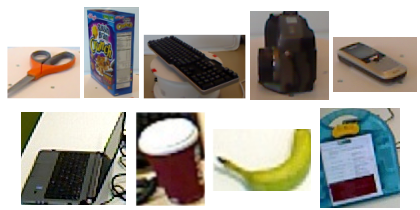}
\caption{First row: examples of images from the Washington dataset \cite{washington}. 
Second row: examples of images of \emph{Dataset A} from \cite{Young2017}.}
\label{fig:wash+strands}\vspace{-3mm}
\end{figure}

\textbf{Feature Extraction.}
We conducted experiments on the Washington RGB-D object dataset \cite{washington}.
This is a widely used testbed in the robotic community, consisting of 41,877 RGB-D images 
organized into 300 instances divided in 51 classes
of common indoor objects (\eg scissors, cereal box, keyboard etc. see top row of Figure \ref{fig:wash+strands}). 
Each object instance was positioned on a turntable and captured from three different 
viewpoints while rotating. Since two consecutive views are extremely similar,
only 1 frame out of 5 is used for evaluation purposes. We did not consider the depth information and 
focused only on the RGB images:  following  
\cite{Carlucci}, we represented the images with the last fully convolutional layer of
the AlexNet network trained on the \emph{IN} dataset and learned a linear SVM model on them. Note that the results obtained with this protocol can be considered the off-the-shelf state of the art in the robot vision community.
Finally we ran a test on the splits provided with the dataset and compared the results
with those obtained by using AlexNet trained on \emph{WINC}. 
We also repeated the evaluation when using a pre-trained GoogLeNet for both \emph{IN} and \emph{WINC},
 considering their last inception layer as features\footnote{Multiple validation experiments have shown
that, when using GoogLeNet, the \emph{inception\_5b} layer produces better results than the following pooling and softmax layers.}.
Results shown in Table \ref{table:washington} indicate that, regardless of the considered network,
the average classification accuracy obtained with \emph{WINC} models is just slightly worse than that obtained with \emph{IN} models 
and they can be considered equal within their standard deviations. It is worthwhile to remember that the Washington 
database was collected in 2011, so with respect to age it can be considered closer to the \emph{IN} dataset than to \emph{WINC}. 
Nevertheless the \emph{WINC} model is robust enough to provide good results when used for feature extraction.

\begin{table}[tb]
\centering
\caption{SVM classification accuracy on the Washington database \cite{washington}. 
The first column indicates the training data and network architecture used to extract the image features.}
\vspace{-2mm}
\begin{tabular}{cc}
\hline
model & classification accuracy (\%)\\
\hline
IN AlexNet & $88.3 \pm 1.9$\\
WINC AlexNet & $87.3  \pm 1.6$\\
IN GoogLeNet & $88.5 \pm 2.3$\\
WINC GoogLeNet & $85.4 \pm 2.5$\\
\hline
\end{tabular}\vspace{-3mm}
\label{table:washington}
\end{table}
\begin{figure}[t]
\centering
\begin{tabular}{@{}cc}
\includegraphics[width=0.5\linewidth]{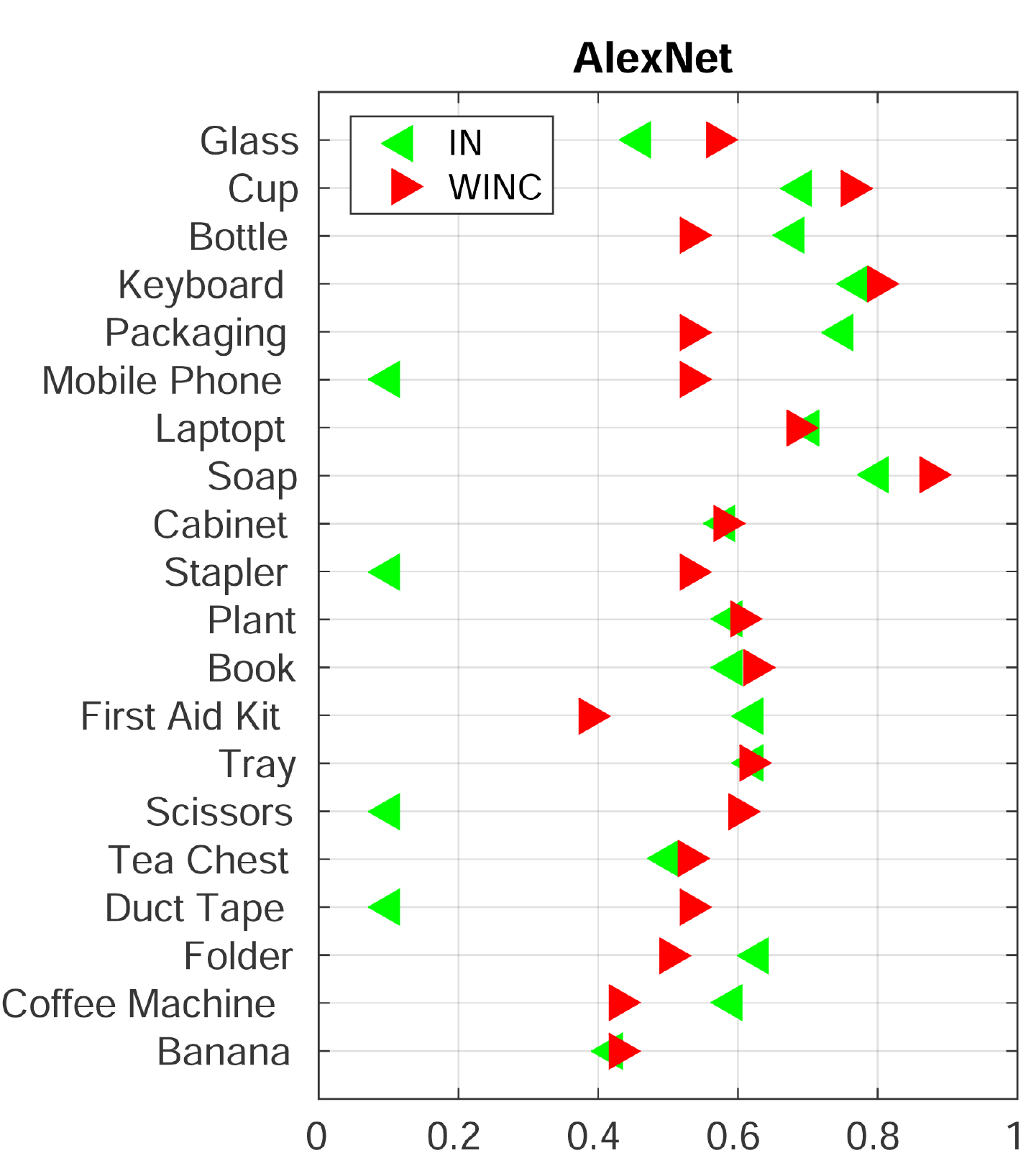} &
\hspace{-2mm}\includegraphics[width=0.355\linewidth]{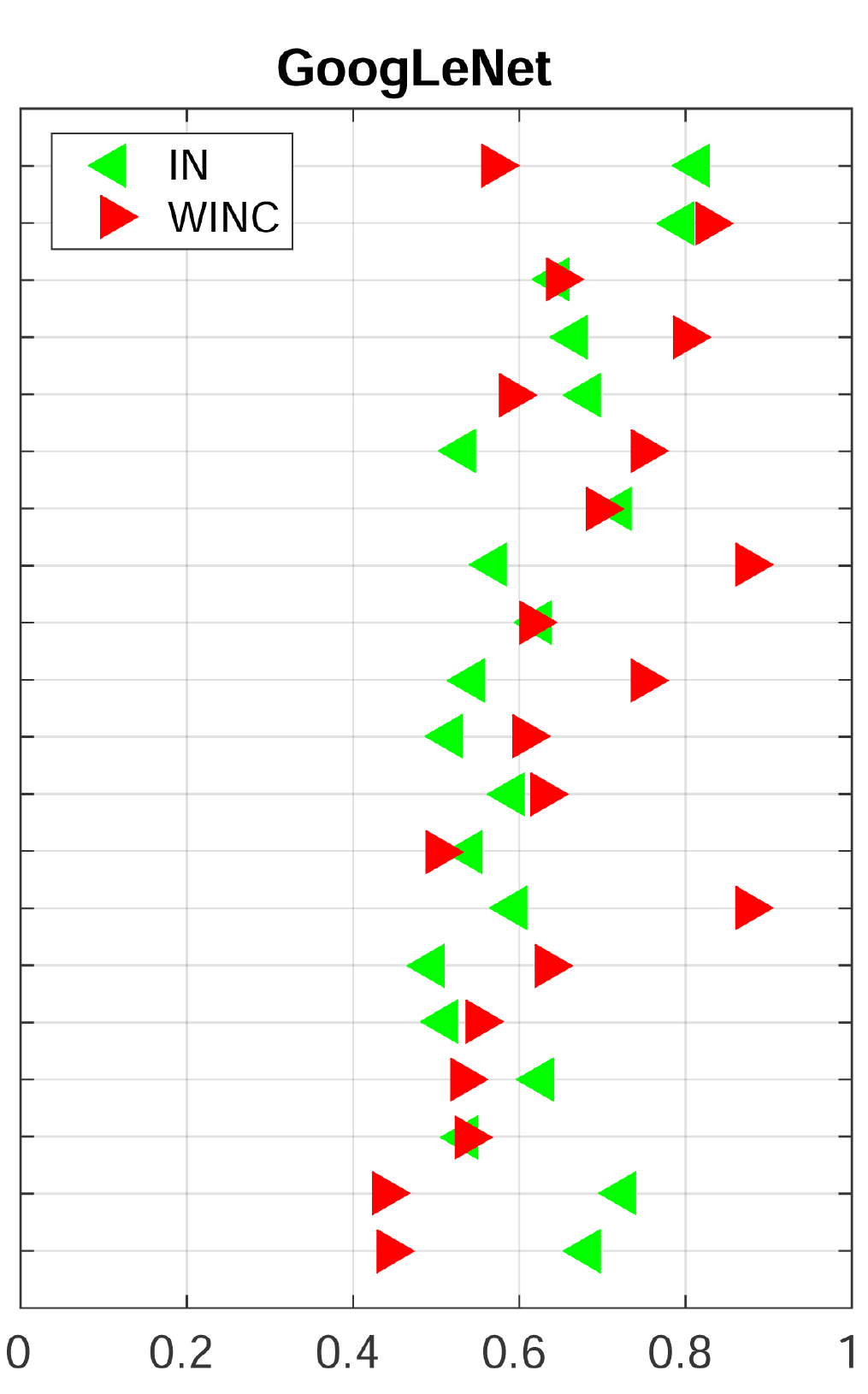}
\end{tabular}
\caption{WUP values between ground truth and predicted object labels. 
The title of each figure indicates the CNN architecture adopted to learn the visual 
classification models which generate the category proposals, while the legend specifies
their training data.
}\vspace{-3mm}
\label{fig:bham}
\end{figure}

\textbf{Category Proposals.}
For a mobile service robot in human environment it is important not only to recognize known 
object categories, but also to be able to generate hypotheses about previously unseen categories. CNN models pre-trained on thousand of
categories can be used as a potential source of information to extend a robot's situated
knowledge. For this task we considered the same setting described in \cite{Young2017}: a long-term autonomous mobile
service robot equipped with an RGB-D camera recording everyday scenes in a human environment~\cite{Hawes:2016}. It 
observes cabinet tops, counters, desks, shelves and other surfaces taking multiple views 
from various angles. From these surfaces it detects and segments objects which it isn't able to recognise~\cite{Faeulhammer:2016}. 
During the test phase, a cropped image of the unknown object is given
as input to a pre-trained CNN model.  Although the correct object label might not be in the predicted set, 
the success is measured by evaluating the WUP similarity \cite{WuPalmer} between the ground truth and 
the proposed category with values ranging in $\{0,1\}$. 

As we deal with autonomously gathered images in real environments they appear smaller, with
lower-resolution and from less favourable angles than both the images in ImageNet and in
our Web-collected dataset (see bottom row of Figure \ref{fig:wash+strands}). Nevertheless, note that using a deep network 
trained on a database created automatically from the Web is a strong proxi for autonomous lifelong 
object discovery from Web images. 

We run the experiments on \emph{Dataset A} from \cite{Young2017}, starting 
from AlexNet models trained both on \emph{IN} and \emph{WINC}. 
Figure \ref{fig:bham} shows the results per each of 20 object categories of the dataset and indicate that 
in most of the cases the WUP score between the ground truth and predicted labels is higher when using 
\emph{WINC} than when using \emph{IN}. On average we get a WUP score of 0.52 with \emph{IN} and 0.58 with \emph{WINC}.
The observed performance is also confirmed by the results obtained when using the GoogLeNet 
architecture with average WUP score of respectively 0.62 and 0.64.
We see these results as a first, important step towards robots able to learn the visual appearance of
previously unseen objects autonomously from the Web.